# Benchmarking Constraint-Based Bayesian Structure Learning Algorithms: Role of Network Topology

Radha Nagarajan[*]
*Children's Hospital of Orange County, California, USA*

Marco Scutari
*Istituto Dalle Molle di Studi sull'Intelligenza Artificiale (IDSIA), Lugano, Switzerland*

**Abstract**

Modeling the associations between real world entities from their multivariate cross-sectional profiles can provide cues into the concerted working of these entities as a system. Several techniques have been proposed for deciphering these associations including constraint-based Bayesian structure learning (BSL) algorithms that model them as directed acyclic graphs. Benchmarking these algorithms have typically focused on assessing the variation in performance measures such as sensitivity as a function of the dimensionality represented by the number of nodes in the DAG, and sample size. The present study elucidates the importance of network topology in benchmarking exercises. More specifically, it investigates variations in sensitivity across distinct network topologies while constraining the nodes, edges, and sample-size to be identical, eliminating these as potential confounders. Sensitivity of three popular constraint-based BSL algorithms (*Peter-Clarke, Grow-Shrink, Incremental Association Markov Blanket*) in learning the network structure from multivariate cross-sectional profiles sampled from network models with *sub-linear, linear,* and *super-linear* DAG topologies generated using preferential attachment is investigated. Results across linear and nonlinear models revealed statistically significant ($\alpha = 0.05$) decrease in sensitivity estimates from sub-linear to super-linear topology constitutively across the three algorithms. These results are demonstrated on networks with nodes ($N_{nods} = 48, 64$), noise strengths ($\sigma = 3, 6$) and sample size ($N = 2^{10}$). The findings elucidate the importance of accommodating the network topology in constraint-based BSL benchmarking exercises.

*Radha Nagarajan, Ph.D.
Children's Hospital of Orange County
1201 W La Veta Avenue, Orange, CA, USA
Email: Radha.Nagarajan@choc.org

**Introduction**
Real-world entities work in concert as a system and respond to environmental cues. While there has been significant progress in capturing high-throughput observational data across real-world entities across diverse paradigms [1], there is considerable interest in deciphering their associations using novel approaches. Such a data-driven approach can assist in hypothesis testing by validating known associations as well as hypothesis generation by discovering novel associations, while providing novel system-level insights, a precursor to evidence-generation and developing effective interventions. Several techniques have been proposed in the literature including *Bayesian Structure Learning* (**BSL**) algorithms for modeling associations from real world observational data [2-5]. BSL models the joint probability distribution of the entities of interest as a product of marginal conditional probability distributions. The resulting network structure is represented as a *Directed Acyclic Graph* (**DAG**) where the nodes and edges represent the entities of interest and their associations respectively. Since BSL algorithms can return DAGs that are probabilistically indistinguishable or Markov equivalent, the resulting structure is represented as a *Partially Directed Acyclic Graph* (**PDAG**) [2, 6]. Under certain implicit assumptions [3], PDAGs may provide cues on potential causal relationships between the entities.

BSL is a NP-hard problem with the number of possible DAG structures increasing super-exponentially with the number of nodes [7, 8]. Several computational approaches have been proposed for learning the DAG structure from multivariate cross-sectional profiles with applications to diverse areas spanning molecular profiling [9, 10], genetics [11], microbiome [12], and healthcare [13]. Computational approaches for BSL broadly fall under *constraint-based*, *score-based*, and *hybrid* approaches [14, 15]. Constraint-based approaches recover the optimal DAG using tests for conditional independence whereas score-based approaches identify the DAG that best represents the given data using a search-criteria in conjunction with a scoring function. Hybrid approaches adopt a combination of constraint-based and score-based approaches in identifying the optimal DAG by minimizing the space of potential DAGs. Recent studies on benchmarking BSL algorithms have focused primarily on the sample size and the dimensionality of the data represented by the number of nodes in the DAG [16-18]. A recent study [19] compared constraint-based, score-based and hybrid approaches, across small and large sample-sizes, as well as across categorical and continuous variables. The results did not reveal a systematic difference in the ranking of the three approaches in terms of speed and accuracy. Subsequent investigation by increasing the sample size improved the goodness of fit and decreased the variability of the learned DAGs. However, the performance was found to plateau at sample sizes around five times the number of parameters of the generating models. The present study is significant as it investigates the impact of network topology on BSL algorithms, while constraining the dimensionality (i.e. nodes, edges) and sample-size of the data to be identical across these topologies. Constraining these parameters to be identical essentially eliminates them as potential contributors to the observed differences in benchmarking measures such as sensitivity. Sensitivity estimates of three popular constraint-based BSL algorithms (**PC**: *Peter-Clarke*, **GS**: *Grow-Shrink*, **IM**: *fast-Incremental Association Markov Blanket*) [20-23] in inferring associations from linear and nonlinear models of DAG structures with markedly different topologies (**B**: *sub-linear*, **L**: *linear*, **U**: *super-linear*) generated using the *preferential attachment* (**PA**) [24, 25] is investigated **Fig. 1**. For the linear model, the signal at a node is determined as the superposition of the aggregate signal from its parent nodes and noise. For the nonlinear model, the signal at a node is determined as the superposition of the nonlinear transform of the aggregate signal from those of its parents and noise. The nonlinear transform was chosen as the static, invertible *sigmoidal* function. The choice of the sigmoidal function can be attributed to its prevalence across DAGs such as feed-forward neural networks [26, 27], and its relevance in modeling real-world phenomenon including (*i*) all-or-none response [28], (*ii*) enzyme kinetics [29], and (*iii*) transcriptional cooperativity [30]. The PA algorithm generates networks whose growth pattern and likelihood of connection $\pi(k)$ scales with the degree centrality ($k$) as $\pi(k) \sim k^\gamma$, where $\gamma$ is the *scaling exponent* [25]. PA relies on the underlying assumption that new nodes prefer to connect to highly connected nodes in a network and accompanied by positively skewed degree centrality distributions observed widely in real-world networks. More specifically, PA overcomes some of the limitations of traditional random graph models such as Erdos-Renyi random graphs [31], accompanied by binomial degree centrality distributions that do not sufficiently capture the growth process of real-world networks. The network growth process in PA can be linear, sub-linear or super-linear, based on the magnitude of the scaling exponent $\gamma$, **Fig. 1**. While linear PA is generated with $\gamma = 1$, sublinear and super-linear PA are generated with $0 < \gamma < 1$ and $\gamma > 1$ respectively. Linear PA is accompanied by a scale-free topology and power-law degree centrality distribution [25]. Sub-linear PA is accompanied by networks with smaller hubs and a stretched exponential degree centrality distribution, whereas super-linear PA is accompanied by hub and spoke topology with a few highly connected nodes [25], **Fig. 1**.

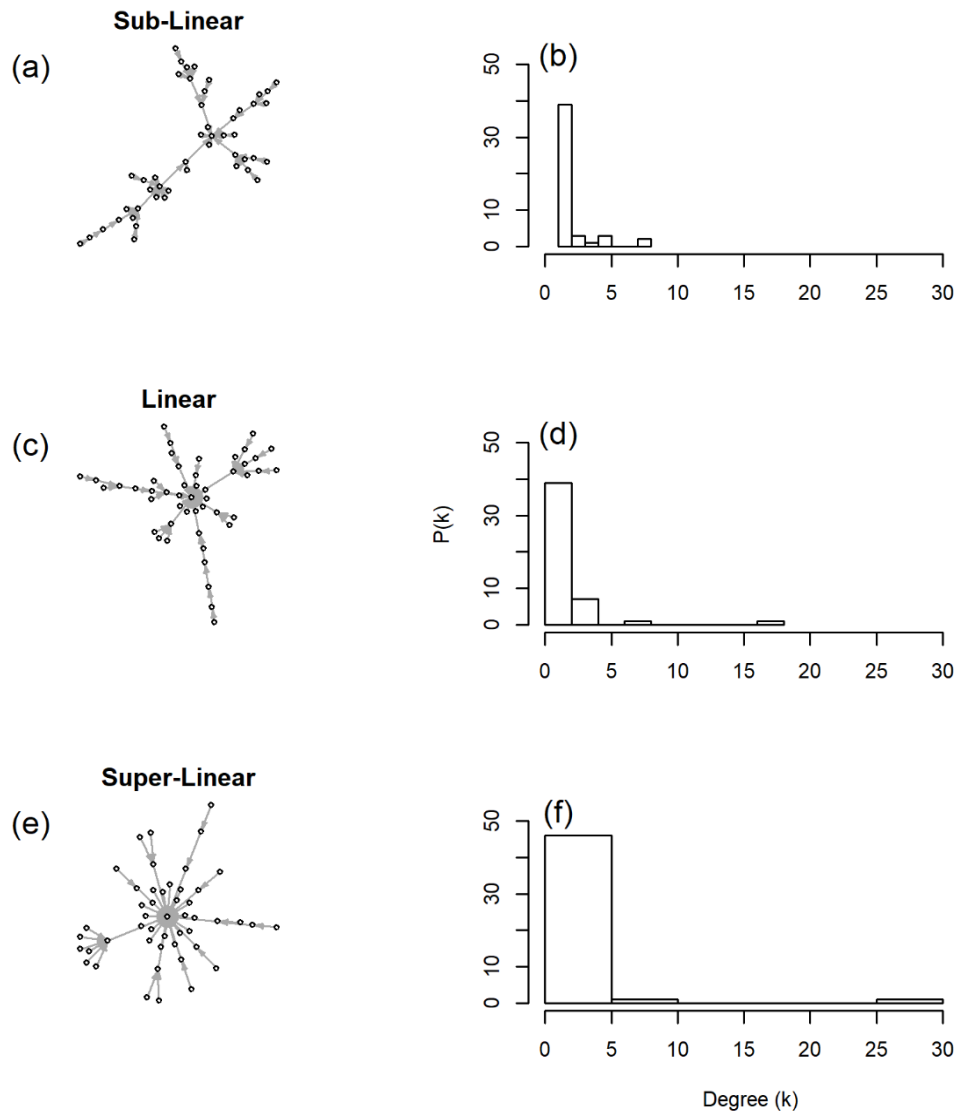

**Figure 1** Representative DAGs with sub-linear ($\gamma = 0.25$), linear ($\gamma = 1.0$), and super-linear ($\gamma = 1.25$) topologies with identical nodes ($N_{nods} = 48$) and edges ($N_{edgs} = 47$) generated using preferential attachment is shown in (a), (c) and (e) respectively. The corresponding in-degree distribution is shown in (b), (d), and (f) respectively.

**Methods**
*Generating Multivariate Cross-sectional Profiles*
Multivariate cross-sectional data were sampled from DAGs with sub-linear, linear, and super-linear topologies and scaling parameters $\gamma = 0.25$, $\gamma = 1.0$ and $\gamma = 1.25$ respectively, **Fig. 1**, using linear and nonlinear models as described below.
- Root nodes in the model represent *signal nodes* (i.e. *in-degree* = 0) and were generated as zero-mean unit variance independent and identically distributed (iid) Gaussian process.
$$x_i = \xi_i \text{ such that } E(x_i, x_j) = 0 \text{ for } i \neq j$$

- Child nodes $x_i$ were generated as the superposition of the signal from its parents $\pi_i$ and zero-mean iid Gaussian noise $\eta_i$ with variance $\sigma^2$ given by the expressions below.
    - *Linear Model*: $x_i = \sum_{j \in \pi_i} \omega_j x_j + \sigma \eta_i$ such that $E(\eta_i, \eta_j) = 0$ for $i \neq j$
    - *Nonlinear Model*: $x_i = \psi(\sum_{j \in \pi_i} \omega_j x_j) + \sigma \eta_i$ where the nonlinear transfer function $\psi$ is given by $\psi(z) = 1/(1 + e^{-z})$

In the above expressions, the coupling strengths $\omega_j$ between the nodes were set to one for convenience.

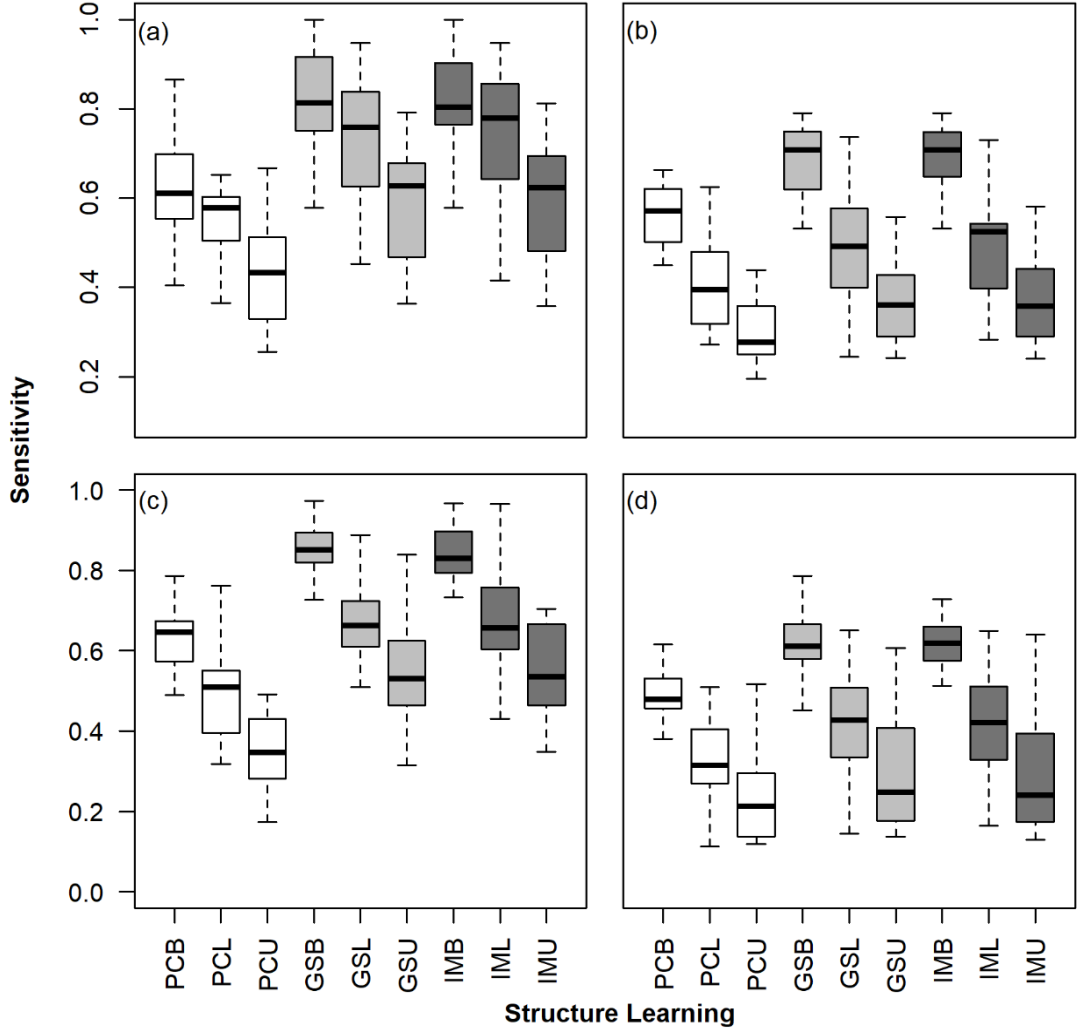

**Figure 2** *Linear Model*: Box-whisker plots of the sensitivity estimates $(N_{reps} = 20)$ of the linear model across the three different network topologies (B, L, U) and the three BSL algorithms PC (white), GS (light gray), IM (dark gray) is shown in the panels (a-d) respectively. Suffixes B, L and U in the panels correspond to sub-linear ($\gamma = 0.25$), linear ($\gamma = 1.0$), and super-linear ($\gamma = 1.25$) topologies. Panels (a) and (b) represent the sensitivity profiles for networks ($N_{nods} = 48, N = 2^{10}$) with noise strengths ($\sigma = 3$) and ($\sigma = 6$) respectively. Panels (c) and (d) represent the sensitivity profiles for networks ($N_{nods} = 64, N = 2^{10}$) with noise strengths ($\sigma = 3$) and ($\sigma = 6$) respectively.

*Benchmarking BSL performance*
The number of nodes ($N_{nods}$), edges ($N_{nods} - 1$), and sample size ($N$) across the DAG topologies were constrained to be identical eliminating these as potential confounders in benchmarking BSL performance. Since the number of edges of the DAG ($N_{nods} - 1$) in this study was significantly lower than the maximum possible edges i.e. $\frac{N_{nods}(N_{nods}-1)}{2}$, sensitivity as opposed to specificity was chosen for benchmarking BSL performance. Sensitivity estimates were determined as the proportion of true edges of the corresponding PDAG generated by moralizing the DAG returned by the BSL algorithm [2]. Sensitivity of the PDAGs were estimated from ($N_{rep} = 20$) independent realizations as a function of increasing noise variance ($\sigma = 3, 6$), number of nodes ($N_{nods} = 48, 64$), for a given sample size ($N = 2^{10}$). Non-parametric statistical tests (Wilcoxon-Ranksum) was used subsequently to assess significant ($\alpha = 0.05$) differences in sensitivity estimates between sub-linear and super-linear topologies. Three BSL algorithms (PC, GS, IM) [21-23] were investigated with default parameters implemented as a part of the R package *bnlearn* [32]. Test for conditional independence for the linear model was chosen as Fisher's Z test, whereas for the nonlinear model was chosen as mutual information with asymptotic $\chi^2$ test and Type I error rate ($\alpha = 0.05$) [32].

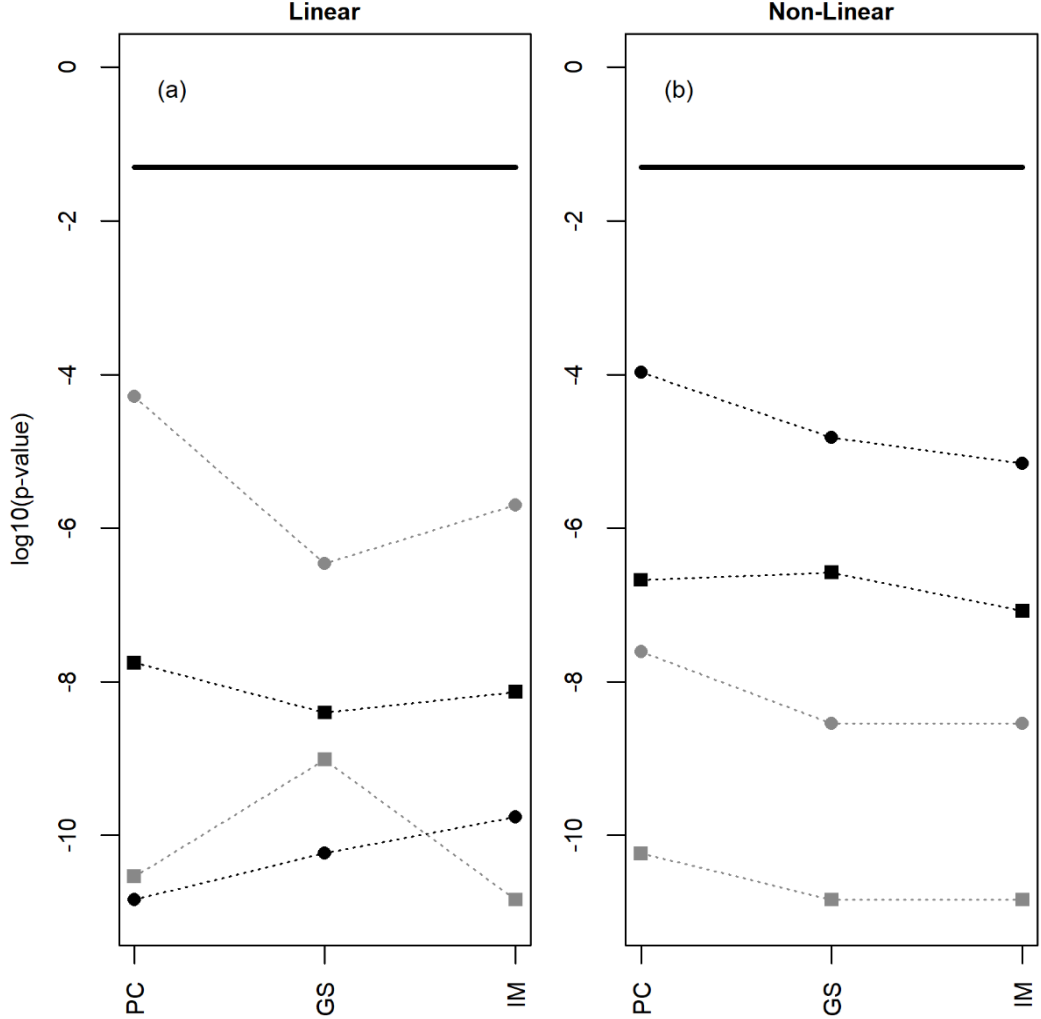

**Figure 3.** p-values from statistical comparison of sensitivity estimates between B and U topologies using Wilcoxon-Ranksum across the three BSL algorithms (PC, GS, IM) with nodes $N_{nods} = 48$ (circles) and $N_{nods} = 64$ (squares), for noise strengths $\sigma = 3$ (gray) and $\sigma = 6$ (black) and sample-size ($N = 2^{10}$), for the linear and nonlinear models is shown in (a) and (b) respectively. The solid horizontal black line corresponds to the significance level ($\alpha = 0.05$) in log-scale and shown as a reference in (a) and (b).

**Results**
Constraint-based BSL as noted earlier uses tests for conditional independence for structure learning and rely on the *local Markov property*, where a given node in the DAG is independent of its non-descendants given its parents. This property especially emphasizes the importance of the in-degree distribution, **Fig. 1**. More importantly, the differences in the in-degree of a node can impact the number of parameters associated with it and performance of the constraint-based BSL algorithm. While the number of nodes and edges (dimensionality) were constrained to be identical across the three topologies in the present study, there were marked differences in the in-degree distributions between the DAG topologies. For instance, the maximum in-degree of the DAGs for sub-linear, linear, and super-linear topologies with dimensionality ($N_{nods} = 48$) were 8, 17, and 28, corresponding to ~17%, ~36% and ~60% of their total edges respectively, **Fig. 1**. Of interest is to that the maximum in-degree of the DAG with super-linear topology was markedly higher than that of sub-linear topology.

*Linear Model*: Sensitivity estimates of the three BSL algorithms (PC, GS, IM) for the linear model, across the three network topologies (B, L, U), with ($N = 2^{10}$) is shown in **Fig. 2**. Sensitivity estimates of (PC, GS, IM) exhibited a decreasing trend with increasing scaling exponent ($\gamma = 0.25, 1.0, 1.25$) for the three topologies (B, L, U). The decrease was statistically significant ($\alpha = 0.05$) between DAG topologies B and U as revealed by Wilcoxon-Ranksum test, **Fig. 3**. This behavior was consistently observed with variation in the dimensionality ($N_{nods} = 48, 64$), noise strengths ($\sigma = 3, 6$) and across the three BSL algorithms. As expected, the sensitivity estimates also exhibited a decrease with increasing noise strengths ($\sigma = 3$ to $6$) and increasing dimensionality ($N_{nods} = 48$ to $64$).

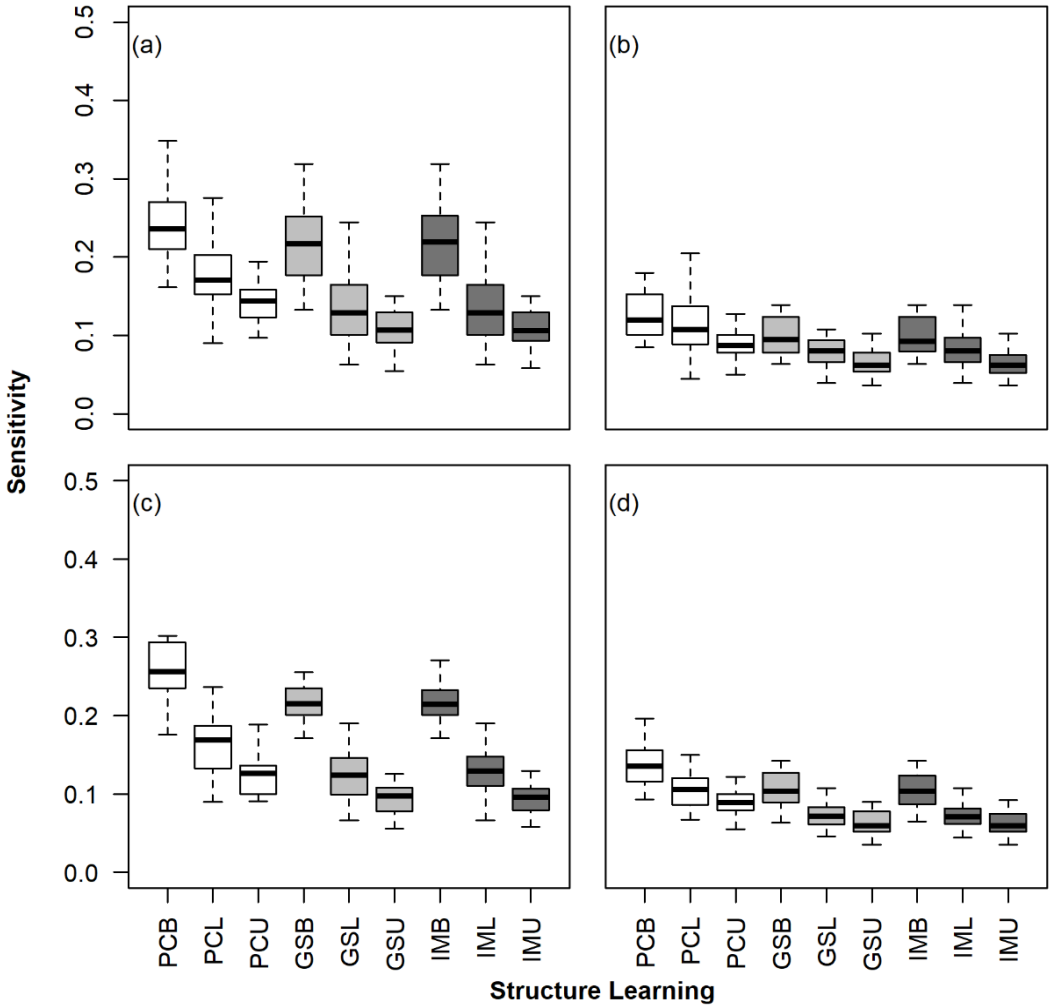

**Figure 4** *Non-linear Model*: Box-whisker plots of the sensitivity estimates ($N_{reps} = 20$) of the nonlinear model across the three different network topologies (B, L, U) and the three BSL algorithms PC (white), GS (light gray), IM (dark gray) is shown in each of the panels (a-d) respectively. For each of the BSL algorithm, the suffixes B, L and U represent sub-linear ($\gamma = 0.25$), linear ($\gamma = 1.0$), and super-linear ($\gamma = 1.25$) topologies. Panels (a) and (b) represent the sensitivity profiles for networks ($N_{nods} = 48, N = 2^{10}$) generated with noise strengths ($\sigma = 3$) and ($\sigma = 6$) respectively. Panels (c) and (d) represent the sensitivity profiles for networks ($N_{nods} = 64, N = 2^{10}$) generated with noise strengths ($\sigma = 3$) and ($\sigma = 6$) respectively.

*Non-linear Model*: Sensitivity estimates of the three BSL algorithms (PC, GS, IM) for the nonlinear model, across the three network topologies (B, L, U), with ($N = 2^{10}$) is shown in **Fig. 4**. While the magnitude of the sensitivity estimates of the nonlinear model was relatively lesser than that of the linear model, the decreasing trend across the three topologies (B, L, U), and statistically significant decrease ($\alpha = 0.05$) between DAG topologies B and U revealed by Wilcoxon-Ranksum test, **Fig. 3** across the three BSL algorithms were similar to that of the linear model. Sensitivity estimates of (PC, GS, IM) exhibited a decreasing trend with increasing scaling exponent ($\gamma = 0.25, 1.0, 1.25$) for the three topologies (B, L, U). This behavior was consistently observed with variation in the dimensionality ($N_{nods} = 48, 64$), noise strengths ($\sigma = 3, 6$) and across the three BSL algorithms. As expected, the sensitivity estimates also exhibited a decrease with increasing noise strengths ($\sigma = 3$ to $6$) and with increasing dimensionality ($N_{nods} = 48$ to $64$).

**Discussion**

While there has been significant progress in generating multivariate cross-sectional data across real-world entities, there is a need for novel approaches in deciphering their associations. Such an understanding is critical for gaining system-level insights and behavior that may not be readily apparent from reductionist representations. Techniques such as constraint-based BSL algorithms can provide system-level insights by modeling the associations as DAGs. These approaches can validate known associations while discovering novel associations from multivariate cross-sectional data. Traditional benchmarking of BSL algorithms had focussed primarily on the dimensionality and sample-size. The present study elucidated marked variations in sensitivity estimates of constraint-based BSL algorithms across distinct network topologies while constraining the dimensionality (i.e. nodes, edges), and sample size to be identical between them. The results were demonstrated across linear and nonlinear models of networks with varying number of nodes, noise strengths, and three popular constraint-based BSL algorithms. More importantly, PDAG sensitivity estimates of the super-linear topologies was shown to be significantly lesser than that of sub-linear topologies across linear and nonlinear models. Variations in the in-degree distributions between the super-linear and sub-linear topology may explain the difference in the sensitivity estimates, since constraint-based BSL rely on the local Markov property. These preliminary findings clearly demonstrate the importance of accommodating network topology in BSL benchmarking exercises in addition to factors such as dimensionality and sample size.

**Reference**
1. Acosta, J.N., et al., *Multimodal biomedical AI.* Nature Medicine, 2022. **28**(9): p. 1773-1784.
2. Koller, D. and N. Friedman, *Probabilistic graphical models: principles and techniques.* 2009: MIT press.
3. Pearl, J., *Causality.* 2009: Cambridge university press.
4. Heckerman, D., *A tutorial on learning with Bayesian networks.* Learning in graphical models, 1998: p. 301-354.
5. Neapolitan, R.E., *Learning bayesian networks.* Vol. 38. 2004: Pearson Prentice Hall Upper Saddle River.
6. Chickering, D.M., *Learning equivalence classes of Bayesian-network structures.* The Journal of Machine Learning Research, 2002. **2**: p. 445-498.
7. Bouckaert, R.R., *Bayesian belief networks: from construction to inference.* 1995.
8. Chickering, M., D. Heckerman, and C. Meek, *Large-sample learning of Bayesian networks is NP-hard.* Journal of Machine Learning Research, 2004. **5**: p. 1287-1330.
9. Sachs, K., et al., *Causal protein-signaling networks derived from multiparameter single-cell data.* Science, 2005. **308**(5721): p. 523-529.
10. Pe'er, D., *Bayesian network analysis of signaling networks: a primer.* Science's STKE, 2005. **2005**(281): p. pl4-pl4.
11. Scutari, M., et al., *Multiple quantitative trait analysis using Bayesian networks.* Genetics, 2014. **198**(1): p. 129-137.

12. Nagarajan, R., et al., *Cross-talk between clinical and host-response parameters of periodontitis in smokers.* Journal of periodontal research, 2017. **52**(3): p. 342-352.
13. Lucas, P.J., L.C. Van der Gaag, and A. Abu-Hanna, *Bayesian networks in biomedicine and health-care.* 2004, Elsevier. p. 201-214.
14. Nagarajan, R., M. Scutari, and S. Lèbre, *Bayesian networks in r.* Springer, 2013. **122**: p. 125-127.
15. Scutari, M. and J.-B. Denis, *Bayesian networks: with examples in R.* 2021: Chapman and Hall/CRC.
16. Kalisch, M. and P. Bühlman, *Estimating high-dimensional directed acyclic graphs with the PC-algorithm.* Journal of Machine Learning Research, 2007. **8**(3).
17. Zuk, O., S. Margel, and E. Domany, *On the number of samples needed to learn the correct structure of a Bayesian network.* arXiv preprint arXiv:1206.6862, 2012.
18. Scutari, M., C. Vitolo, and A. Tucker, *Learning Bayesian networks from big data with greedy search: computational complexity and efficient implementation.* Statistics and Computing, 2019. **29**: p. 1095-1108.
19. Scutari, M., C.E. Graafland, and J.M. Gutiérrez, *Who learns better Bayesian network structures: Accuracy and speed of structure learning algorithms.* International Journal of Approximate Reasoning, 2019. **115**: p. 235-253.
20. Spirtes, P., C. Glymour, and R. Scheines, *Causation, prediction, and search.* 2001: MIT press.
21. Colombo, D. and M.H. Maathuis, *Order-independent constraint-based causal structure learning.* J. Mach. Learn. Res., 2014. **15**(1): p. 3741-3782.
22. Margaritis, D., *Learning Bayesian network model structure from data.* 2003, School of Computer Science, Carnegie Mellon University Pittsburgh, PA, USA.
23. Tsamardinos, I., et al. *Algorithms for large scale Markov blanket discovery.* in *FLAIRS.* 2003.
24. Barabási, A.-L. and R. Albert, *Emergence of scaling in random networks.* science, 1999. **286**(5439): p. 509-512.
25. Pósfai, M. and A.-L. Barabási, *Network science.* 2016: Citeseer.
26. Macintyre, A. and E.D. Sontag. *Finiteness results for sigmoidal "neural" networks.* in *Proceedings of the twenty-fifth annual ACM symposium on Theory of computing.* 1993.
27. Rosenblatt, F., *The perceptron: a probabilistic model for information storage and organization in the brain.* Psychological review, 1958. **65**(6): p. 386.
28. Dayan, P. and L.F. Abbott, *Theoretical neuroscience: computational and mathematical modeling of neural systems.* 2005: MIT press.
29. Goldbeter, A. and D.E. Koshland Jr, *An amplified sensitivity arising from covalent modification in biological systems.* Proceedings of the National Academy of Sciences, 1981. **78**(11): p. 6840-6844.
30. Veitia, R.A., *A sigmoidal transcriptional response: cooperativity, synergy and dosage effects.* Biological Reviews, 2003. **78**(1): p. 149-170.
31. Bollobás, B. and B. Bollobás, *Random graphs.* 1998: Springer.
32. Scutari, M., *Learning Bayesian networks with the bnlearn R package.* arXiv preprint arXiv:0908.3817, 2009.